\documentclass[sigplan, nonacm]{acmart}
\AtBeginDocument{%
  }

\setcopyright{acmlicensed}
\copyrightyear{2018}
\acmYear{2018}
\acmDOI{XXXXXXX.XXXXXXX}
\acmConference[EuroSys'26]{EuroSys: European Conference on Computer Systems}{April 13--16,
  2026}{Edinburgh, Scotland}

\acmSubmissionID{183}

\usepackage{tikz}
\usepackage{amsmath}

\usepackage{algorithm}
\usepackage{algorithmic}
\usepackage{multirow}
\usepackage{subfigure}
\usepackage{booktabs} 

\begin{document}

\title{PRESERVE: Prefetching Model Weights and KV-Cache in Distributed LLM Serving}

\author{Ahmet Caner Y\"{u}z\"{u}g\"{u}ler}
\email{ahmet.yuzuguler@huawei.com}
\affiliation{%
  \institution{Huawei Zurich Research Center}
  \country{Switzerland}
}

\author{Jiawei Zhuang}
\email{zhuangjiawei@hisilicon.com}
\affiliation{%
  \institution{Huawei Zurich Research Center}
  \country{Switzerland}
}

\author{Lukas Cavigelli}
\email{lukas.cavigelli@huawei.com}
\affiliation{%
  \institution{Huawei Zurich Research Center}
  \country{Switzerland}
}

\renewcommand{\shortauthors}{Y\"{u}z\"{u}g\"{u}ler et al.}

\begin{abstract}
Large language models (LLMs) are typically served from clusters of GPUs/NPUs that consist of large number of devices. Unfortunately, communication between these devices incurs significant overhead, increasing the inference latency and cost while limiting the scalability. Prior work addressed this issue by overlapping communication with compute, but has severe limitations due to the data dependencies between these operations. In this paper, we propose PRESERVE, a novel framework that prefetches model weights and KV-cache from off-chip HBM memory to the on-chip cache of AI accelerators during the communication operations, which offers various advantages and performance improvements compared to prior methods. 

Through extensive experiments conducted on commercial AI accelerators, we demonstrate up to 1.6× end-to-end speedup on state-of-the-art, open-source LLMs. Additionally, we perform a design space exploration that identifies the optimal hardware configuration for the proposed method, showing a further 1.25× improvement in performance per cost by selecting the optimal L2 cache size. Our results show that PRESERVE has the potential to mitigate the memory bottlenecks and communication overheads, offering a solution to improve the performance and scalability of the LLM inference systems.

\end{abstract}

\ccsdesc[500]{Do Not Use This Code~Generate the Correct Terms for Your Paper}
\ccsdesc[300]{Do Not Use This Code~Generate the Correct Terms for Your Paper}
\ccsdesc{Do Not Use This Code~Generate the Correct Terms for Your Paper}
\ccsdesc[100]{Do Not Use This Code~Generate the Correct Terms for Your Paper}

\keywords{LLM inference, distributed systems, compute-communication overlap, L2 prefetching}


\maketitle

\section{Introduction}

Large language models (LLM) have been widely deployed across various services and application domains such as chat assistants~\cite{OpenAI23}, code generation~\cite{Chen21}, and knowledge retrieval~\cite{Lewis20}. While LLMs have demonstrated outstanding capabilities in these domains, their substantial computational requirements result in slow and costly inference. To address these challenges, specialized accelerators have been developed and deployed for LLM inference. Consequently, the performance, efficiency, and scalability of LLM inference accelerators are critical to the speed and cost of the LLM inference.

\begin{figure*}[t]
    \centering
    \includegraphics[width=1.\textwidth]{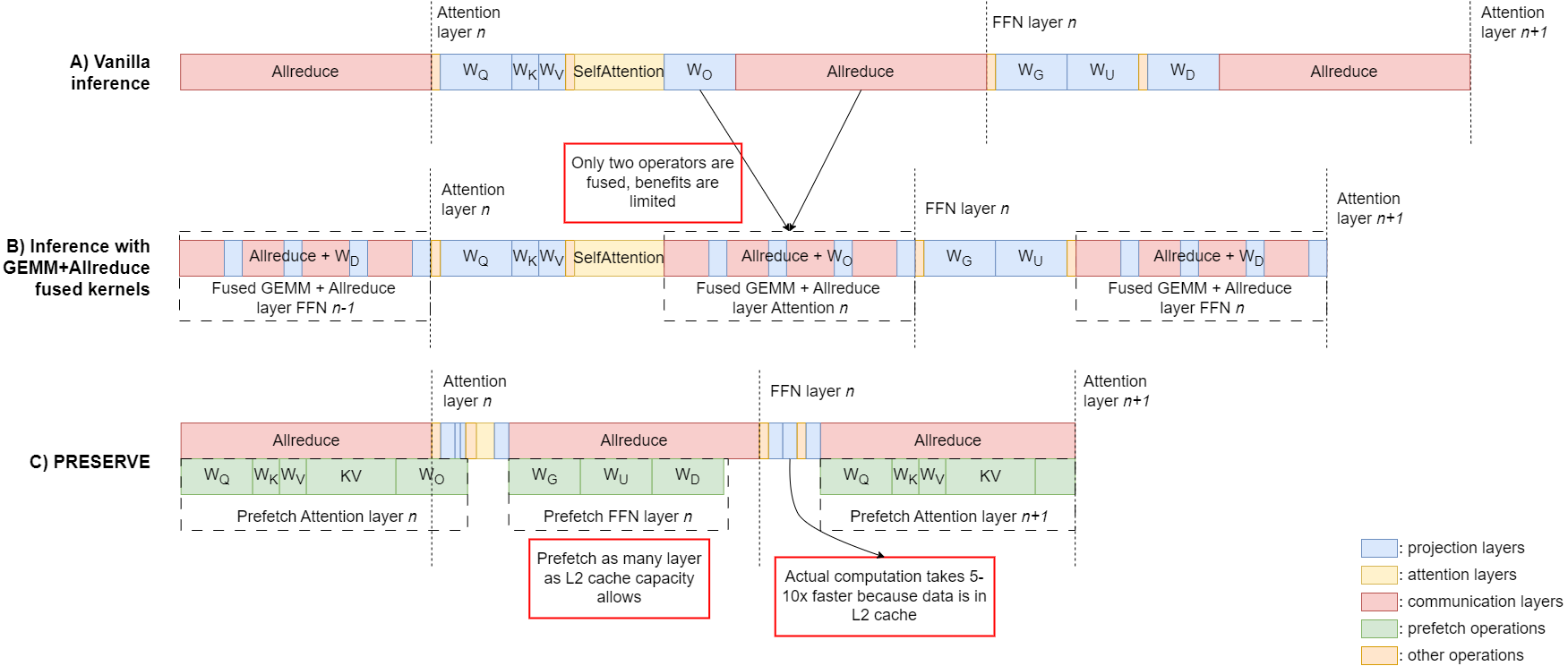}
    \caption{Conceptual comparison between A) vanilla inference, B) inference with GEMM+Allreduce fused kernels~\cite{Rashidi21, Hoefler08, Punniyamurthy24, Chang24, WangWei23}, C) PRESERVE (this work).}
    \label{fig:preserve-comparison}
\end{figure*}

One of the most significant computational challenges in today's LLM inference systems is the off-chip memory bottleneck. While the prefill phase of LLM inference allows all input tokens to be efficiently processed at once, the output tokens in the decoding phase are typically generated one by one due to the autoregressive nature of LLMs. Therefore, the model weights and contextual data (also known as KV-cache) must be retrieved from off-chip memory during each iteration of the decoding phase. Unfortunately, even the most advanced memory technologies such as HBM fail to provide sufficient bandwidth to overcome these memory bottleneck issues. As a result, the performance of the decoding phase, and consequently the overall LLM inference, is often limited by the HBM bandwidth of the accelerators.

Another significant computational challenge in LLM inference is the large memory footprint of the model weights and KV-cache. The total memory size of model weights and KV-cache often exceeds the HBM capacity of a single accelerator. Therefore, LLM inference is typically performed on clusters of accelerators ranging from a few up to hundreds of devices. During such distributed inference scenarios, the accelerators must share intermediate results with each other between layers using collective communication operations such as \textit{allreduce}. During these communication phases, the accelerators mostly remain idle, leading to reduced device utilization.

To mitigate the communication overhead, prior studies have proposed fusing matmul and allreduce operations to overlap compute and communication \cite{Rashidi21, Hoefler08, Punniyamurthy24, Chang24, WangWei23}, as shown in Figure \ref{fig:preserve-comparison}. While these methods might improve the resource utilization by overlapping one GEMM and one consecutive allreduce operations, their effectiveness is limited as they can only fuse two consecutive operations, which is often insufficient to hide the entire communication latency in large inference clusters. Moreover, these methods can not be applied to KV cache, because there are many other operations between self-attention and allreduce operations in a typical LLM architecture. Finally, operator fusion requires high engineering efforts and expertise in low-level kernel development, which also reduces the practicality of these methods. As a result, existing compute-communication overlap methods fail to provide an efficient, effective and comprehensive solution.

In this paper, we address the communication overhead in distributed LLM inference by exploiting the ever-growing on-chip memory capacity in modern AI accelerators (e.g., NVIDIA GB200: 126 MB of L2 cache~\cite{gb200}, AMD MI300X: 256 MB of L3 cache~\cite{mi300x}, Huawei Ascend 910B: 196 MB of L2 cache~\cite{ascend}). We propose prefetching the model weights and KV-cache from off-chip HBM memory to on-chip cache of the AI accelerators during the communication operations to overlap memory reads for the model weights and KV-cache with communication, hiding the latency of the latter. Once the communication operations are completed, compute cores access the required data from the on-chip cache, no longer limited by off-chip memory bandwidth. As a result, the overall decoding phase of LLM inference is executed faster and more efficiently, reducing end-to-end latency and execution cost.


Due to the data dependencies between the communication operations and the subsequent matmul operations, neither hardware schedulers nor modern ML compiler can automatically and trivially perform prefetching on model weights and KV-cache. Moreover, if not handled carefully, prefetching excessive amounts of data poses the risk of polluting the cache, which may result in a slowdown rather than a speedup. Therefore, we propose \textbf{PRESERVE}, a novel method that prefetches model weights and KV-cache from off-chip HBM memory to the on-chip device cache in parallel to the collective communication operations. To maximize the performance gains without the need for any modification in the user code for ease of development, we implemented PRESERVE as a graph optimization framework. Our graph optimization scheme inserts \textit{prefetch} instructions into a given computation graph, to be executed in parallel streams to the collective communication operations. PRESERVE also keeps track of the prefetched data and estimates L2 cache usage during compilation time in a way that the total prefetched data size at any time does not exceed a predefined upper limit in order to prevent cache pollution and potential slowdowns. 

To demonstrate the performance gains of PRESERVE, we first conduct a series of experiments on a commercial AI accelerator and show up to 1.6$\times$ end-to-end speedup on open-source state-of-the-art LLMs. To identify the optimal hardware design for an AI accelerator using the proposed weight and KV-cache prefetching method, we perform a design space exploration. The results of our design space exploration show that the optimal L2 cache size increases from 8 MB to 104 MB when prefetching is taken into account, leading to a further 1.25$\times$ improvement in performance per cost compared to the baseline accelerator design.

In short, this paper makes the following contributions:

\begin{itemize}
    \item We introduce PRESERVE, a novel prefetching framework for model weights and KV-cache in LLM inference that overlaps HBM reads with collective communication operations to hide their latencies.
    \item We propose a graph optimization scheme that automatically and optimally inserts prefetching operations into the computational graph of LLM inference.
    \item We perform a series of experiments on commercial AI accelerators, demonstrating up to 1.6$\times$ end-to-end speedup on open-source state-of-the-art LLMs. 
    \item We conduct a design space exploration to identify the optimal hardware configuration with the proposed method in place, showing a further 1.25$\times$ improvement in performance per cost when the optimal L2 cache size is used.
\end{itemize}

The rest of this paper is organized as follows: Section 2 provides background information on LLM inference and AI accelerators. Section 3 introduces the proposed method and details of the proposed graph optimization scheme. Section 4 explains our experiment methodology and presents the experimental results. Section 5 elaborates on our design space exploration and discusses our findings. Section 6 reviews the prior work and highlights the novelty of this work. Finally, Section 7 concludes the paper and discusses the key takeaway messages.

\section{Background}

Since their inception, the Transformer-based LLM models have been widely in use in many application fields. However, their inference is typically overwhelmingly costly, which impedes their widespread adoption. In this section, we give a brief overview of these architectures and their computational characteristics to understand the limitations on the performance of their inference. 

\paragraph{LLM inference workloads}
Most of today's state-of-the-art generative AI LLM models (e.g., GPT-4~\cite{OpenAI23} and Llama-2~\cite{Touvron23}) consist of a \textit{decoder-only Transformer architecture}~\cite{Vaswani17}. The decoder-only Transformer models comprise a number of \textit{decoder} layers stacked on top of each other. Each decoder layer includes a self-attention and a multi-layer perceptron (MLP) layer, each of which are followed by normalization layers. The self-attention layers can further be decomposed into three input projection layers, a positional embedding, a self-attention kernel (e.g., Flash attention~\cite{Dao22}), and an output projection layer. Similarly, the MLP layers can be decomposed into three projection layers and one activation layer. A large percentage of memory footprint and computational workload of Transformer architectures comes from the projection layers and self-attention kernels, which puts them at the center of focus for LLM acceleration.

The generation process with LLMs can be divided into two phases: \textit{prefill} and \textit{decode}. In the prefill phase, the LLMs process the input prompts and generate the first output token. In the decode phase, the LLMs keep generating one token at a time based on the past tokens due to their autoregressive nature, until a special end-of-sequence token is generated or the maximum number of token limit is reached. Because the length of input prompts can easily exceed hundreds or thousands of tokens in most generative AI applications, the matrix multiplication operations of linear layers and self-attention kernels in a typical prefill phase greatly benefit from data reuse, which increases their operational intensity (in terms of FLOPS/bytes) and maximizes the utilization of the compute resources.

\paragraph{LLM decoding is memory BW-limited}
In contrast, the algebraic operations in self-attention layers during the decode phase are mostly matrix-matrix mutliplication with one very narrow matrix (often no wider than 8), thus providing an operational intensity of ~16 Op/word while most accelerators' roofline is at >100 Op/word. Similarly, the data reuse in projection layers is limited to the number of batches in the decode phase, which also often result in low operational intensity. Therefore, in various application scenarios where the number of batches might be limited to a small number due to low rate of incoming queries or limited memory capacity (e.g., mobile platforms or latency-constraint online services), the decode phase suffers from memory-bandwidth bottlenecks. As a result, the autoregressive generative AI applications with LLMs is generally considered a memory-bandwidth bottlenecked process. 

To eliminate redundant computation in autoregressive steps of the decode phase, the calculated key and value tensors are stored in memory to be reused in the later iterations. This mechanism is generally referred to as \textit{KV-cache}. In generation tasks with long context lengths (e.g., book summarization or multi-modal models), the size of KV-cache might easily become comparable or exceed the size of model weights. Moreover, reading KV-caches from off-chip memory might take even longer than the model weights for long context lengths. Thus, improving the performance of handling and reading KV-caches is critical to the overall performance of LLMs. As a result, in this work, we do not only consider prefetching for model weights but also for KV-caches.

\begin{figure*}[h!]
    \centering
    \includegraphics[width=1.\textwidth]{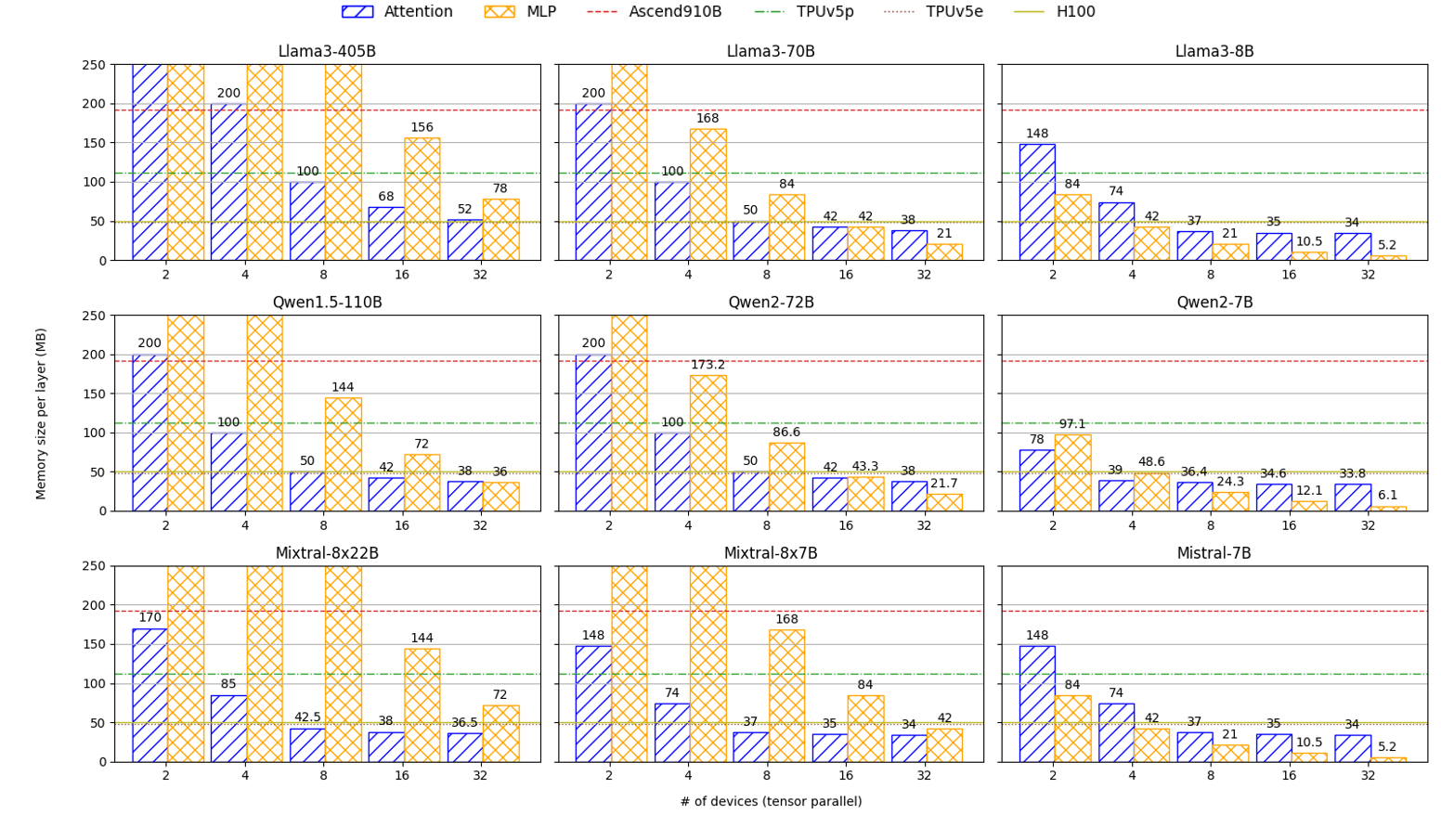}
    \caption{Memory footprint of Attention and MLP layers of various LLMs for varying number of devices. The models are assumed to be distributed using tensor parallelism, and both weights and KV-cache are stored in int8 precision. Batch size and context length are taken as 8 and 16k, respectively. Horizontal lines represent the L2 capacity of various state-of-the-art NPU and GPUs, namely Ascend~910B~\cite{ascend}, TPUv5p~\cite{Vahdat2023tpuv5p}, TPUv5e~\cite{Vahdat2023tpuv5e}, and H100~\cite{h100}.}
    \label{fig:l2-req}
\end{figure*}

\paragraph{Multi-device inference}
Many of today's state-of-the-art LLMs have model sizes that reach hundreds of billions of parameters. Even with reduced precision formats (e.g., bfloat16, float8, int8), the memory footprint of these models exceed the HBM memory capacity of individual modern AI accelerator devices. Moreover, the throughput of a single device is typically bounded by the HBM bandwidth, which is in the order of a few TB/s for high-end AI accelerators. With today's state-of-the-art models exceeding hundreds of GBs in model size and KV-cache, a single device is not sufficient to achieve per-token latency objectives that are required to ensure a responsive user experience, which is often set to 100 ms per token.
Therefore, the LLM inference is typically distributed and executed in parallel across multiple tightly-interconnected devices.

Various parallelization strategies have been proposed and used in distributed LLM inference to minimize the overhead of communication between devices. One of the most commonly used parallelization strategy is called \textit{tensor}-parallelism. In this parallelization technique, the weights and the KV-cache in LLMs are partitioned and distributed across the devices. During inference, each device performs computation locally with their partitions of weights and KV-cache, and sums up their results through the communication primitive \textit{allreduce}. 

To minimize the communication overhead, the self-attention layers are typically partitioned in the \textit{attention-head} dimension, where the calculation of each attention head is independent of each other up to the output projection layer of the self-attention layers~\cite{Shoeybi19}. Similarly, the MLP layers are also partitioned in their \textit{intermediate} dimension, which corresponds to the columns and rows of the input and output linear projections, respectively, so that much of the calculation can be performed locally. As a result of this partitioning scheme, accelerators need to perform only one allreduce call per self-attention and MLP layer.

Even though only a single allreduce call per layer is needed for distributed inference, their execution may take considerable time depending on the data and cluster size. The data size that needs to be communicated across the accelerators grows linearly with the embedding size of the LLM architecture and the batch size of the input prompts. Moreover, the network traffic also grows linearly with the number of accelerators participating in the tensor parallel execution. While modern LLM inference servers are equipped with high-bandwidth interconnection fabric between the devices (e.g., NVLink and HCCS), scaling beyond a single server still requires performing allreduce calls over slower networks such as PCIe or InfiniBand. As a result, the execution time of allreduce calls may take a large portion of the overall time and become the limiting factor in the scalability of the distributed LLM inference systems.

In summary, there are two distinct computational characteristics of today's LLMs that limit the performance of their inference. First, the underlying operations exhibit low operation intensity, which makes the single-device performance of LLM inference limited by the memory bandwidth. Second, LLMs must be distributed across multiple devices due to their large memory footprint and latency constraints, which incurs significant communication overheads. These two computational challenges limit the performance and efficiency of the AI inference systems, increasing the cost per token and hindering their scalability and responsiveness, particularly in real-time applications.


\section{Proposed Method}
\label{sec:method}

In the previous section, we argued that memory-bandwidth bottlenecks and communication overheads are the two most significant obstacles to the scalability and performance in distributed LLM inference systems. For this reason, in this section, we propose a method that prefetches the model weights and KV-cache from HBM memory to L2 cache and overlaps prefetching with communication in order to hide the latency of communication with memory operations.

\paragraph{Vanilla execution}
In vanilla execution, the query, key, and value linear projections of the attention layers have a data dependency with the preceding allreduce operation, as they need to wait for the reduction of the activations from the previous layer. Similarly, the gate and up projections of the MLP layers also have a data dependency on the preceding allreduce operation. As a result, modern ML frameworks (e.g., Pytorch) execute these operations sequentially in a single stream. Unfortunately, the devices remain idle while waiting for allreduce operations, leading to underutilization and long latencies. 

\paragraph{Execution with the proposed method}
While the linear projections need to wait for allreduce operations due to data dependencies on activations, their weights are read-only, so they do not have any dependencies with the preceding operations. Similarly, the $K$ and $V$ caches in the $QK^T$ and $SV$ calculations of the self-attention layers are also read-only in the decode stage, except for the last entry, which is updated based on the key and value calculated in the preceding linear projections. Therefore, the weights and the KV caches can be prefetched before performing the forward pass of these layers. 

With the proposed method, devices start prefetching the weights and KV-cache in a parallel stream from off-chip HBM memory to on-chip buffers (e.g., L2 cache) while waiting for the allreduce operations. When the allreduce operation is completed and the device is ready to execute the next layers, the weights and KV-cache are already fetched to the L2 cache, which typically provides 5-10$\times$ higher bandwidth than off-chip HBM memories. As a result, the forward pass of the linear projections and the self-attention takes much shorter, improving resource utilization and speeding up the decode phase. 

\paragraph{L2 capacity requirements}
The proposed method requires AI accelerators to store the prefetched weights and KV-cache of the subsequent layers in their on-chip memory, which is typically the L2 cache. If the L2 cache size of the accelerators is not sufficient to store the weights and KV-cache, the prefetched data would be evicted and the effectiveness of the proposed method would be reduced. Therefore, the AI accelerators should have enough L2 cache size to store all the weights and KV-cache of the layers between the allreduce operations. 

To assess the feasibility of the proposed method on popular LLMs, we now analyze the memory requirements of the individual LLM layers and compare them against the L2 cache size of modern AI accelerators. Fig. \ref{fig:l2-req} shows the calculated memory size of the Attention and MLP layers of various LLMs for varying number of devices. As we increase the number of devices, provided that the number of devices does not exceed the number of attention heads, the weights and KV-cache are partitioned into smaller chunks, which decreases the memory size per device.

Fig. \ref{fig:l2-req} demonstrates that tensor parallelism up to 16 devices is sufficient to reduce the memory requirements of the Attention and MLP layers of even the largest models down to a level at which they can fit in the L2 cache of the existing commercial AI accelerators. Modern LLM inference systems often consist of many more devices. For instance, Deepspeed-inference scales up to 256 GPUs~\cite{Aminabadi22} and Cloud TPUv5e inference instances contain up to 256 chips~\cite{Vahdat2023tpuv5e}. Therefore, we conclude that the proposed method is applicable to most of the widely used, open-source LLMs given the scale of the current LLM inference systems and the L2 cache capacity of the commercial AI accelerators.

\begin{figure}[t]
\centering     
\includegraphics[width=0.48\textwidth]{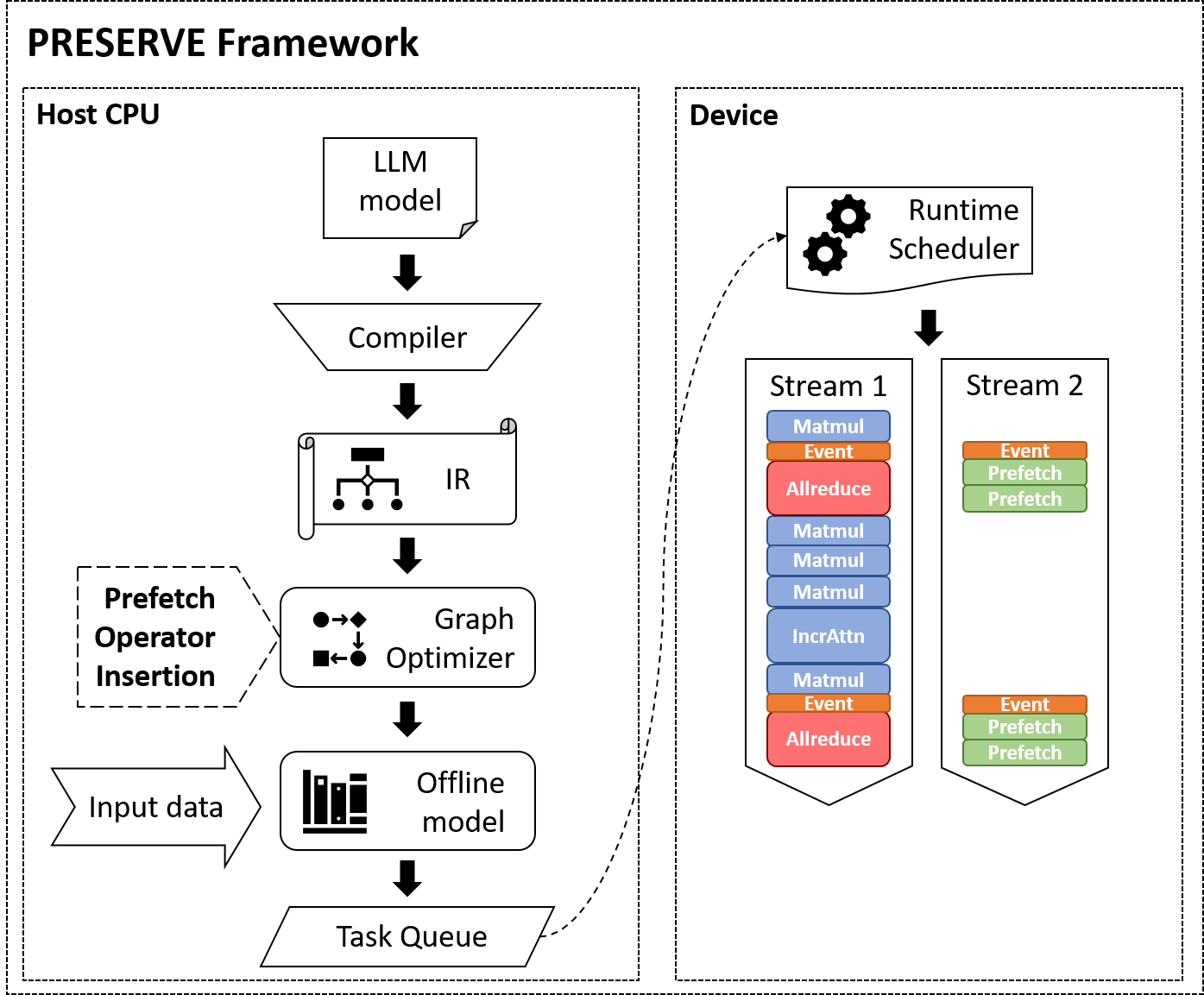}
\caption{An overview of the proposed PRESERVE framework.}
\label{fig:preserve-overview}
\end{figure}

\paragraph{Framework integration}
How the proposed method is exposed to the user is critical for its effectiveness and applicability. A trivial implementation of the proposed method would be to develop a prefetch operation as a custom operator and offload the task of inserting prefetch operators to the programmers. However, this approach would have three shortcomings. First, the programmers would need to explicitly insert the prefetch operators in parallel to the communication operations, which would increase the burden of the programmers. Second, the programmers would need to have a good understanding of the memory hierarchy of the underlying hardware platforms to realize the full potential of the proposed method. Third, the programs written in this way would be specific to certain hardware platforms and could not be easily migrated to different ones without performance degradation. Therefore, the proposed method does not expose prefetching to the programmers but automatically inserts the required operators in the computation graph of an ML application.

Many of today's hardware vendors provide graph optimization frameworks (e.g., CUDA Graphs~\cite{cudagraph}, CANN Graph Engine~\cite{cann} etc.), which take a program written in a hardware-agnostic high-level programming language (e.g., Pytorch) and perform compiler-level optimizations, such as operator insertion, fusion, removal. Moreover, such graph optimization frameworks typically have the full view of the entire graph, which enables to optimize for best use of memory resources. To that end, we also developed the PRESERVE framework, which inserts prefetching operators for weights and KV-caches into a computational graph of LLM inference, while automatically managing stream synchronization and cache optimizations in order to minimize cache pollution and redundant memory bandwidth usage.

\paragraph{PRESERVE}
Figure \ref{fig:preserve-overview} illustrates the overview of the proposed PRESERVE framework. On the host side, the framework takes a user code that implements an LLM model inference written in one of the supported ML frameworks such as Pytorch and Tensorflow. Then, the user code is compiled into an intermediate representation (IR) using either the built-in compiler of the ML frameworks (e.g., TorchDynamo) or one of the hardware-specific compilers (e.g., Ascend ATC~\cite{atc}). Next, IR is passed to a graph optimizer (e.g., Ascend Graph Engine (GE)~\cite{Liao19}), which performs graph-level optimizations and inserts prefetch operators for the suitable compute and communication operations into the graph. Finally, the optimized graph is compiled into an offline model (executable binaries) using a vendor-specific operator library (e.g., Ascend CANN~\cite{cann}).

Upon receiving a request, PRESERVE loads the offline model, copies the input data and model weights to the device memory, and sends the compiled operators to the device using a task queue. When the runtime scheduler in the device receives the compiled operators, it dispatches the operators according to the execution order and stream that they are compiled for. The prefetch operators are executed in a stream parallel to the main stream and synchronized between layers using \textit{event} instructions to facilitate the overlap between communication and prefetching.

\paragraph{Operator insertion}
Algorithm \ref{alg:insert} describes the proposed operator insertion algorithm for weight and KV-cache prefetching. The proposed algorithm takes a computation graph and L2 cache capacity of the target hardware platform as inputs. Then, the algorithm finds and iterates over all communication operations in the computation graph. For each communication operation, the algorithm visits the children nodes in a breadth-first order until it reaches the end of graph or the next communication operation. For each children node of the type \textit{MatMul} or \textit{SelfAttention}, the algorithm estimates the required memory size for prefetching and calculates the total memory size of all preceding prefetch operators between the current node and the communication operation. If the total memory size is smaller than the L2 cache capacity, a prefetch operator for the weight or KV-cache of the current node is inserted in a parallel stream to the communication operation. If the total memory size exceeds the L2 cache capacity, the prefetch operator is not inserted to avoid cache eviction, and the algorithm continues with the next communication operation.

\begin{algorithm}[tb]
   \caption{Prefetch Operator Insertion}
   \label{alg:insert}
\begin{algorithmic}
   \STATE {\bfseries Input:} compute graph $g$, L2 cache size $C$
   \FOR{each comm. op $O$ in $g$}
       \STATE Initialize $reachedEOG = false$.
       \STATE Initialize $cacheSum = 0$.
       \REPEAT
        \STATE Get node $n$ from $O.childNodes()$ using BFS

        \IF{$n$ is a comm. op}
            \STATE Set $reachedEOG = true$
        \ELSIF {$n$ is a type of MatMul or SelfAttention}
            \STATE $cacheSum += n.memSize()$
            \IF {$cacheSum < C$} 
                \STATE Insert prefetch op $P$
            \ELSE
                \STATE Set $reachedEOG = true$
            \ENDIF
        \ENDIF
       \UNTIL{$reachedEOG$ is $true$}
   \ENDFOR
\end{algorithmic}
\end{algorithm}

In conclusion, we propose a method that automatically inserts prefetch operators in parallel to the communication operations in the computation graph of an LLM inference application. In runtime, the prefetch operators enables AI accelerators to start reading the weights and KV-caches while waiting for the communication operations, which hides the latency of the latter and improves the performance of the LLM inference.

\section{Experiments}
\label{sec:experiments}

To evaluate the performance improvements obtained by the proposed method, we performed a series of experiments. In this section, we first explain our experimental methodology, benchmark models, and system details. Then, we test the proposed method under various settings such as tensor-parallel cluster size, batch size, and sequence length. Finally, we discuss our findings and outline the impact of the proposed method on the performance of LLM inference. 

\paragraph{Experimental setup}
For the experiments in this section, we choose a number of open-source LLMs that are widely adopted by the community: Llama3-8b, Llama3-70b~\cite{Touvron23}, Qwen2-7B, Qwen2-72B~\cite{Yang24}, Phi-3-small, and Phi-3-medium \cite{Abdin24}. We varied the batch sizes from 1 to 64 and the sequence length from 2k to 32k to explore the impact of batch size and sequence length on the performance of the proposed method. For sake of simplicity, we assume static batching with equal sequence lengths. The prefill and decode lengths are taken as the 2/3 and 1/3 of the sequence length, respectively.

All benchmarks are implemented in Pytorch with the \textit{torch-npu} backend~\cite{torchnpu}, based on the reference implementation given in \textit{torchair} library~\cite{torchair} and compiled using TorchDynamo. The parallel execution is achieved via the communication primitives provided by the \textit{torch.distributed} library. 
The proposed method is implemented using the Graph Engine (GE) feature of the Compute Architecture for Neural Networks (CANN)~\cite{cann}, which enables to perform weight and KV-cache prefetching during inference.

The experiments are performed on a Huawei Atlas 800T A2 server with 4x Kunpeng~920 48-core CPUs and 8x Ascend~910B NPUs. Each NPU has an L2 cache capacity of 192 MB and is equipped with 24 DaVinci AI cores~\cite{Tang23} and 64 GB HBM memory, which provides a total theoretical FLOPs of 800 TOPS/s in int8 precision and 1.6TB/s off-chip memory bandwidth~\cite{ascend, li2024}. The Ascend~910B NPUs in the server are tightly interconnected to each other through an HCCS interconnect fabric in a full-mesh topology. 

\begin{table}[t]
\caption{End-to-end inference execution time in seconds with the baseline and proposed methods for various benchmarks and number of NPUs. Batch size and maximum sequence length are taken as 4 and 16k, respectively. Both activations and weights are in int8 format. }
\label{tab:main-results}
\vskip 0.15in
\begin{center}
\begin{small}
\begin{tabular}{lccccc}
\toprule
\multirow{2}{*}{Model} &  \multirow{2}{*}{\# NPUs} & \multicolumn{2}{c}{Execution time (s)} & \multirow{2}{*}{Speedup} \\
    & & Baseline & This Work & \\
\midrule
\multirow{ 3}{*}{Llama3-8B } 
    & 2 & 148.8 & 111.3 & 1.34$\times$ \\
    & 4 & 128.2 & 80.6 & 1.59$\times$ \\
    & 8 & 60.3 & 54.5 & 1.11$\times$ \\
\midrule
\multirow{ 3}{*}{Llama3-70B } 
     & 2 & 596.0 & 515.8 & 1.16$\times$ \\
     & 4 & 501.5 & 368.7 & 1.36$\times$ \\
     & 8 & 233.8 & 211.1 & 1.11$\times$ \\
\midrule
\multirow{ 3}{*}{Qwen2-7B } 
     & 2 & 130.9 & 111.6  & 1.17$\times$ \\
     & 4 & 68.8 & 63.0 & 1.09$\times$ \\
     & 8 & 57.6 & 52.7 & 1.09$\times$ \\
\midrule
\multirow{ 3}{*}{Qwen2-72B } 
     & 2 & 700.0 & 616.2 & 1.14$\times$ \\
     & 4 & 593.6 & 438.4 & 1.35$\times$ \\
     & 8 & 402.8 & 342.4 & 1.18$\times$ \\
\midrule
\multirow{ 3}{*}{Phi3-small } 
     & 2 & 147.8 & 110.0 & 1.34$\times$ \\
     & 4 & 128.4 & 79.9 & 1.61$\times$ \\
     & 8 & 59.2 & 54.2 & 1.09$\times$ \\
\midrule
\multirow{ 3}{*}{Phi3-medium } 
     & 2 & 192.8 & 157.3 & 1.23$\times$ \\
     & 4 & 171.8 & 119.9 & 1.43$\times$ \\
     & 8 & 89.3 & 80.4 & 1.11$\times$ \\

\bottomrule
\end{tabular}
\end{small}
\end{center}
\vskip -0.1in
\end{table}

\paragraph{End-to-end execution times}
Table \ref{tab:main-results} summarizes the end-to-end LLM inference execution times with the baseline implementation and the proposed method for various benchmarks and cluster sizes (number of NPUs). The results show that the proposed method speeds up the execution in all benchmarks and cluster sizes. We observe that the speedup generally increases with the number of NPUs. This trend is due to the fact that, as the number of NPUs increases, the computation time per device decreases while the communication time between the devices increases. As a result, hiding the latency of the communication operations has a greater impact on the execution time, which increases the speedup. 

We observe that the maximum speedups occur when the number of NPUs is equal to four (except Qwen2-7B, where it occurs when the number of NPUs is equal to two). This is because the number of KV-heads per device is equal to one when these models are partitioned to eight devices. When the number of KV-heads per device is equal to one, memory read operations for the KV-cache from HBM are continuous rather than strided; thus, the HBM bandwidth is utilized better and memory operations become faster, leaving smaller room for improvement for the proposed prefetching method. Therefore, we observe the highest speedups when the number of local KV-heads per device is equal to two. 

\begin{figure*}[t!]
\centering     
\subfigure[Llama3-70B]{\label{fig:speedup-llama3-70b}\includegraphics[width=0.33\textwidth]{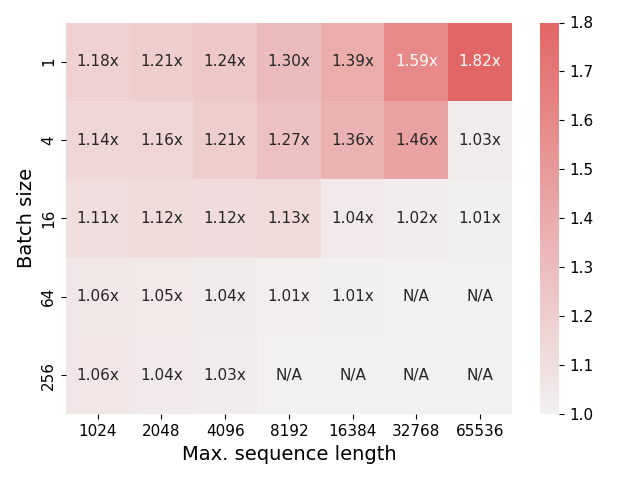}}
\subfigure[Qwen2-72B]{\label{fig:speedup-qwen2-72b}\includegraphics[width=0.33\textwidth]{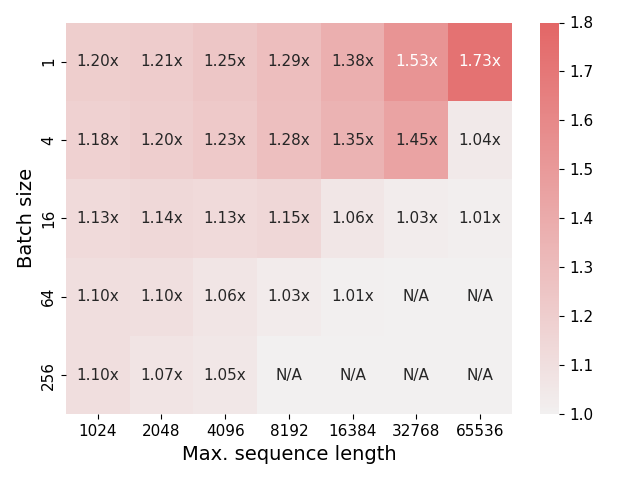}}
\subfigure[Phi3-medium]{\label{fig:speedup-phi3-medium}\includegraphics[width=0.33\textwidth]{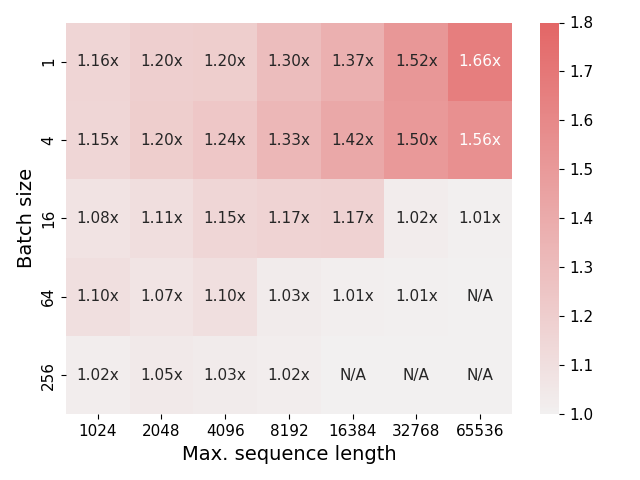}}

\caption{Speedups obtained with the proposed method for varying batch size and maximum sequence length for various models. Both activation and weights are in int8 precision. The experiments are performed on four NPUs. N/A indicates insufficient HBM capacity. }

\label{fig:batch-seqlen-result}
\end{figure*}

We obtain the highest speedups ($\sim$1.6$\times$) with Llama3-8B and Phi3-small models, which are higher than their larger counterparts, namely Llama3-70B and Phi3-medium (1.35$\times$ and 1.43$\times$, respectively). This is due to the fact that the communication to computation ratio is higher in the small models; thus, overlapping memory reads with communications leads to higher speedups. Unlike Llama3-8B and Phi3-small, the maximum speedup with Qwen2-7B is, however, limited to 1.17$\times$. This is because the number of attention heads in Qwen2-7B is smaller than the other models; thus, it does not scale well with increasing number of NPUs. Overall, the experiments demonstrate that the proposed method significantly improves the end-to-end execution time of various state-of-the-art models for varying number of NPUs, with speedups ranging from 1.09$\times$ up to 1.61$\times$.

\paragraph{Batch size \& sequence length}
The effectiveness of the proposed method is sensitive to the batch size and maximum sequence length because of the KV-cache and its implications on the computational characteristics and memory requirements. To that end, we perform a number of experiments to evaluate the impact of batch size and maximum sequence length on the speedup obtained by the proposed method. In these experiments, we run the last iteration of the decode stage for an LLM architecture with the given batch size and sequence length for multiple times with and without the proposed method and calculate the speedup based on the median execution times.

Figure \ref{fig:batch-seqlen-result} shows the speedups obtained with the proposed method for the largest three models used in the previous experiments with batch size and sequence lengths ranging from 1 to 256 and from 1k to 64k, respectively. The results of this experiment show that, for the given range of batch size and sequence length, the proposed method achieves a speedup up to 1.82$\times$. The largest speedups are obtained for long sequence lengths and small batch sizes. The speedup generally decreases with the batch size for two reasons. First, as the batch size increases, the computation in MLP layers becomes less memory-bound, which reduces the effectiveness of the proposed method in MLP layers. Second, the L2 requirement of the KV-cache increases with the batch size. Therefore, for large batch sizes, the KV-cache does not fit in L2, leading to no speedup for the self-attention operators.

We also observe that the speedup generally increases with the sequence length until it reaches a certain threshold. The reason for the increase in the speedup is due to the fact that the speedup obtained in the self-attention layers are higher than the MLP layers due to the formers' irregular and strided memory access patterns. As a result, increasing the sequence length leads to higher end-to-end speedup, as the self-attention layers take up a larger portion of the total execution time with larger sequence lengths. However, the memory requirements of the KV-cache also increases with the sequence length. As a result, increasing the sequence length beyond a threshold where the KV-cache no longer fits in L2 renders the proposed method ineffective for self-attention layers, significantly reducing the speedup. Nevertheless, this drop in the speedup occurs only in very large sequence lengths (e.g., 32k for a batch size of 4), where most practical applications require much smaller sequence lengths. Therefore, the effective range of the proposed method covers a wide range of batch size and sequence lengths (up to a batch size of 256 or a sequence length of 64k).

\paragraph{Comparison to compute-communication overlap fused kernels}
To demonstrate the performance benefits of the proposed method over the prior methods using fused kernels that overlap GEMM and Allreduce operations, we now compare the performance of PRESERVE over a baseline, in which the last linear layer of each Attention and MLP layer, namely $W_{out}$ and $W_{down}$ respectively, are fused with the succeeding Allreduce operations. To fuse the GEMM and Allreduce operations, we use the \textit{npu\_mm\_all\_reduce\_base} \cite{ascend_mc2} operator from the \textit{torch\_npu} library. We then measure the execution time of Llama3-70B (fp16) decoding with both PRESERVE and the baseline, on a cluster of 4x Ascend~910B NPUs for batch sizes ranging from 16 to 1024 and maximum sequence lengths ranging from 128 to 8192, both incrementing as powers of two.   

\begin{figure}[t!]
\centering     
\includegraphics[width=0.4\textwidth]{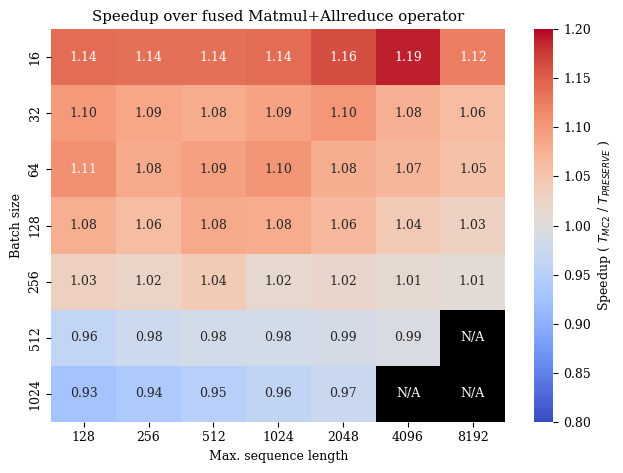}
\caption{Comparison over the fused GEMM+Allreduce operator baseline for various batch size (y-axis) and max. sequence lengths (x-axis) for Llama3-70b in fp16. The red-colored squares indicate that PRESERVE is faster than the fused GEMM+Allreduce. N/A denotes out-of-memory. }
\label{fig:mc2}
\end{figure}

Figure \ref{fig:mc2} shows that PRESERVE outperforms the baseline at small batch sizes while the baseline performs better at large batch sizes. There are multiple reasons why PRESERVE is advantageous over the fused Matmul+Allreduce operators at small batch sizes. To begin with, PRESERVE allows overlapping the HBM reads of any weights and KV-cache with communication, whereas the Matmul+Allreduce operators can only fuse consecutive operations, which limits their applicability only to the last projection layers of each layer, omitting the KV-cache entirely. Moreover, the fused operators typically partition the matmul operations into multiple chunks and pipeline the execution to hide the latency of the communication. However, small chunks of data often lead to inefficiencies in memory accesses, which inhibit the performance of these operators at small batch sizes. As a result, PRESERVE outperforms the baseline with fused Matmul+Allreduce operators at small batch sizes with a speedup up to $1.19\times$.

In contrast, due to higher computational intensity and the growing size of KV-cache (as already discussed before and shown in Figure~\ref{fig:batch-seqlen-result}), the effectiveness of PRESERVE slowly diminishes with the increasing batch size. As a result, the baseline becomes faster than PRESERVE at batch sizes equal to or larger than 512, shown by the less-than-one speedups in Figure \ref{fig:mc2}, with a minimum speedup of $0.93\times$ at the batch size of 1024 and max. sequence length of 128. Therefore, we conclude that PRESERVE is the better choice in scenarios, in which the batch size is smaller than 512 (e.g., online and resource-constraint inference), whereas the fused Matmul+Allreduce operators proposed by the prior work remain as the method of choice for scenarios that require batch sizes equal to or larger than 512 (e.g., offline inference).

In conclusion, in this section, we evaluated the effectiveness and the performance of the proposed method for various LLMs under diverse inference settings, such as the number of NPUs, batch size, and sequence lengths. Our experiments demonstrated that the proposed method achieves up to 1.61$\times$ end-to-end speedup on commercial AI accelerators. Our study on batch size and sequence length identified the optimal settings where the proposed method is most effective and showed that the proposed method achieves significant speedups, outperforming the baseline methods under a wide range of inference settings.

\section{Design Space Exploration}

The experiment results in the previous section demonstrate significant improvement in the performance of distributed LLM inference with PRESERVE framework on existing commercial AI accelerators. However, various hardware design parameters and specifications have an impact on the performance of the PRESERVE framework. Therefore, in this section, we investigate how the hardware design parameters affect the effectiveness of the PRESERVE framework and perform a design space exploration to identify the optimal hardware design parameters that maximize the inference performance of AI accelerators.

\paragraph{Performance \& cost model}
To perform the design space exploration presented in this section, we developed a performance model of AI accelerators for Transformer architectures. We modeled the compute throughput as a linear function of the \textit{flops} capacity of an accelerator. We assume that the memory read and write throughputs are linear functions of the off-chip memory or on-chip L2 cache bandwidths, in case of a L2 miss or hit, respectively. We modeled the communication latency as the summation of a constant initial latency and the data size divided by the bandwidth of the link between the accelerators. We assume that multiple accelerators in a cluster are interconnected in a ring topology.

\begin{table}
\centering
\caption{Model parameters used in design space exploration. }
\label{tab:dse-parameters}
\begin{tabular}{lr}
\toprule
Tech node                                     & 7\,nm                           \\
Area per core                                 & 1.34\,mm\textsuperscript{2}     \\
Area per 1 MB L2 SRAM                         & 0.36\,mm\textsuperscript{2}     \\
Throughput per core (int8)                    & 1.84\,TeraOps/s                 \\
Power consumption per core                    & 0.526\,Watts                    \\
Throughput                                    & 800\,TeraOps/s                  \\
HBM capacity                                  & 64\,GB                          \\
HBM bandwidth                                 & 1.84\,TB/s                      \\
L2 bus bandwidth                              & 12\,TB/s                        \\
Interconnect bandwidth                        & 200\,Gbit/s                     \\
Interconnect latency                          & 25\,\textmu s                   \\ 
\bottomrule
\end{tabular}
\end{table}

Table \ref{tab:dse-parameters} summarizes the values of the model parameters used in our design space exploration. 
The silicon area for cores and L2 SRAM cache are taken from Lin et al.~\cite{Lin20}, whose design is fabricated in a 7nm technology node. 
We assumed that the accelerator has a theoretical peak throughput of 800 TeraOps/s, which is similar to existing inference accelerators~\cite{ascend}~\cite{li2024} and it is equipped with 4x HBM2e memory, which provides 64 GB of capacity and 1840 GB/s bandwidth in total~\cite{synopsys_hbm}. 

For the design space exploration, we selected a number of open-source, state-of-the-art LLMs, namely Llama3-8B, Llama3-70B, Llama3-405B~\cite{Touvron23}, Qwen2-7B, Qwen2-72B, Qwen1.5-110B~\cite{Yang24}, Phi3-small, Phi3-medium, Phi3.5-MoE~\cite{Abdin24}, Mistral-7B, Mixtral-8x7B, and Mixtral-8x22B~\cite{Jiang24}. Unless specified otherwise, we assume a cluster size of 32 NPUs and we vary the batch size from 8 to 32 with increments of 8 and maximum sequence length from 2k to 16k with steps that correspond to the powers of 2k. In all experiments, we assume that both activations and weights are in int8 quantization.

\begin{figure}[t!]
\centering     
\includegraphics[width=0.48\textwidth]{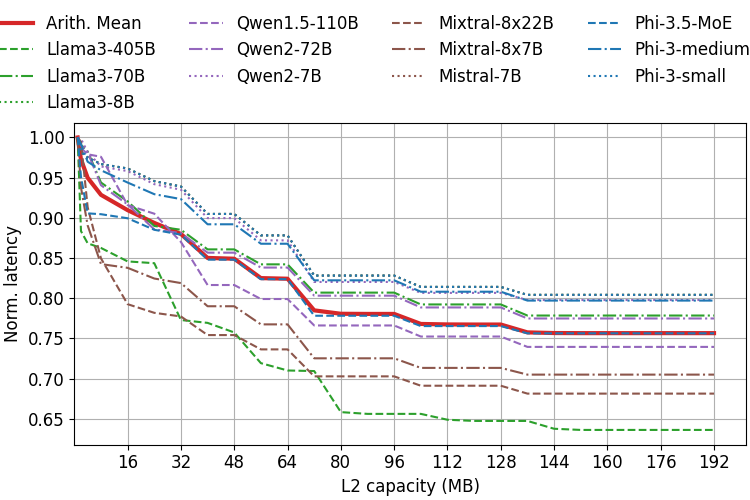}
\caption{Inference latency with respect to L2 cache size for various LLMs and their arithmeric mean, normalized to the smallest L2 cache size.}
\label{fig:lat-vs-l2}
\end{figure}

\paragraph{Performance impact of L2 size}
The size of the L2 cache is critical for the applicability and effectiveness of the proposed prefetching mechanism. Unfortunately, the L2 cache size of today's commercial AI accelerators are provisioned without taking weight and/or KV-cache prefetching into consideration. As a result, such AI accelerators cannot fulfill the full potential of the proposed method due to insufficient L2 capacity. To that end, we first perform a design space exploration to understand the requirements of the L2 cache size for weight and KV-cache prefetching and identify the optimal cache size that maximizes the performance/cost ratio.

To evaluate the impact of L2 cache size on the speedups obtained by the proposed method, we vary the L2 cache size and calculate the latency for various models and benchmarks. Figure \ref{fig:lat-vs-l2} shows the inference latency for each model, normalized to the smallest L2 capacity. We observe that, with increasing L2 cache size, the latency is reduced as more data can be prefetched to L2. We observe that the latency is reduced by percentages ranging between 20\% and 36\%, with an average of 25\%. We observe that the maximum speedups for most models are achieved at the L2 cache sizes of 136 and 144 MB, beyond which we do not observe any further speedups.

\begin{figure}[t!]
\centering     
\includegraphics[width=0.48\textwidth]{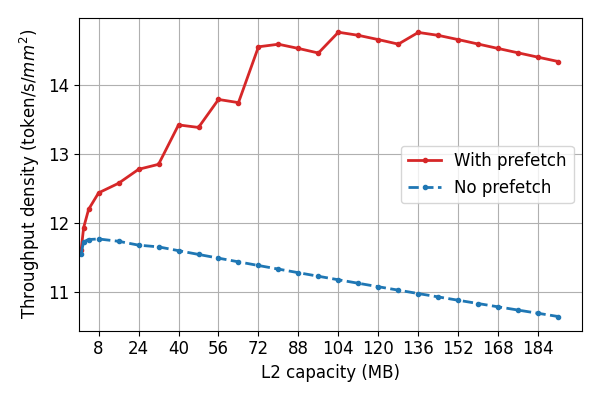}
\caption{Throughput density in terms of token/s/mm$^2$ with respect to L2 cache size, with and without prefetching enabled. The curves have peak throughput density of 14.8 and 11.8 token/s/mm$^2$ at 104 and 8 MB of L2 cache for with and without prefetching, respectively.}
\label{fig:optimal-l2}
\end{figure}

\paragraph{Throughput-area trade-off}
While the effectiveness of the proposed method generally increases with the L2 cache size, allocating a larger L2 cache also increases the silicon area, leading to an accelerator design with a higher cost per unit. Therefore, to identify the sweet-spot between the performance improvement and cost overhead in terms of the L2 cache size, we devise the \textit{throughput density} metric, which is equal to the LLM inference throughput in terms of tokens/s divided by the total die area in terms of mm$^2$. To show the impact of weight and KV-cache prefetching on the throughput density and optimal L2 cache size, we perform a design space exploration with and without prefetching enabled, by sweeping a range of L2 cache size and calculating the throughput density. 

Figure \ref{fig:optimal-l2} illustrates the throughput density with respect to the L2 cache size with and without prefetching enabled. We observe that the throughput density for the baseline accelerator design without prefetching peaks at 8 MB of L2 cache, beyond which it decreases due to the diminishing returns on the benefits of the increasing L2 size. For an accelerator that uses prefetching, on the other hand, the throughput density increases up to the L2 cache size of 104 MB, as more model weights and KV-cache are prefetched and more communication overhead is hidden thanks to the larger L2 cache size. Similar to the baseline, the throughput density decreases beyond the L2 cache size of 104 MB, as the cost overhead of larger L2 cache outweighs the benefits of prefetching.

Our design space exploration shows that the optimal L2 cache size changes for inference accelerators when prefetching is taken into account. Without prefetching, a relatively small L2 cache size (i.e., 8 MB) is sufficient to cache the required data for individual operations in order to minimize off-chip memory accesses. In contrast, an accelerator design for the proposed prefetching scheme requires a larger L2 cache size (i.e., 104 MB) in order to store more model weights and KV-cache during collective communication operations. Nevertheless, prefetching more model weights and KV-cache improves the performance density even at the expense of increased L2 cache size, increasing the  throughput density from 11.8 to 14.8 token/s/mm$^2$, which corresponds to an improvement by 1.25$\times$.


\begin{figure}[t!]
\centering     
\includegraphics[width=0.48\textwidth]{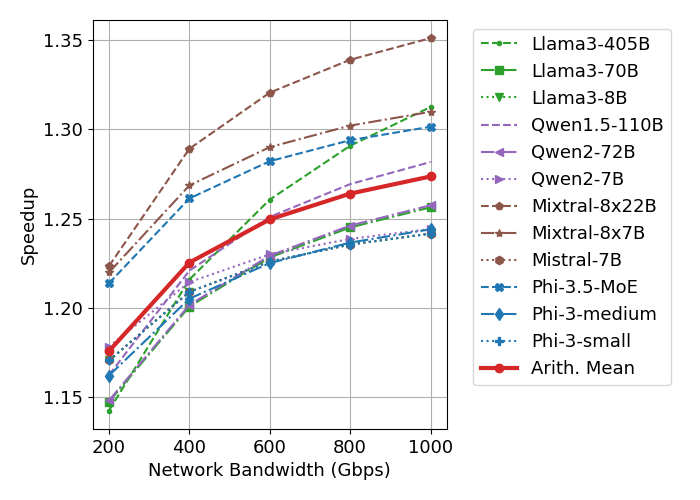}
\caption{Speedup obtained with prefetching over the baseline for various LLMs with respect to the network bandwidth of the links between the devices. The L2 cache size and the number of devices are taken as 104 MB and 128, respectively. The batch size and max. sequence length are equal to 16 and 8k, where prompt and decode lengths are 2/3 and 1/3 of the total number of tokens, respectively. }
\label{fig:bandwidth}
\end{figure}

Besides the L2 cache, the network bandwidth between the devices also has a considerable impact on the speedups achieved by the proposed method, as it directly affects the allreduce latencies. Figure \ref{fig:bandwidth} plots the speedups obtained by the proposed method for various LLMs with respect to the network bandwidth. We observe that the speedup generally increases with the network bandwidth for all models. This is due to the fact that increasing the bandwidth reduces the allreduce latencies, which enables to hide a larger portion of the communication overhead, leading to a higher speedup. On average, we observe that increasing the device-to-device link bandwidth from 200 to 1000 Gbps elevates the speedup from 1.18$\times$ to 1.27$\times$ for a cluster that consists of 128 devices.

In short, we perform a design space exploration using a performance model for AI accelerators to understand the impact of hardware design parameters and specifications on the performance and efficiency of the PRESERVE framework. We show that the optimal L2 cache size increases from 8 MB to 104 MB when prefetching is taken into account while designing the AI accelerators, which improves the throughput density by 1.25$\times$ even in spite of the increased silicon area due to larger L2 cache. 

\section{Related Work}

The large memory footprint and high computational cost of LLMs have encouraged researchers to focus on improving the efficiency, scalability, and performance of LLM inference systems. In this section, we present a summary of the most relevant prior art on the topics of efficient kernel implementation, parallelization techniques, system optimizations and software prefetching and we emphasize the novelty of our proposed method over the prior work. 

\paragraph{KV-cache optimization \& operator fusion}
Due to the autoregressive nature of Transformer architectures, the performance of LLM inference is mostly bounded by off-chip memory bandwidth. To mitigate the memory bottlenecks, researchers have proposed several efficient kernel implementations. \textit{Flash Attention} and \textit{Deepspeed-Inference} resort to tiling and operator fusion to minimize the off-chip bandwidth usage and increase the efficiency of self-attention layers~\cite{Dao22, Aminabadi22}. Kwon et al. proposed \textit{PagedAttention}, in which the memory management of KV-caches are optimized to improve the memory usage~\cite{Kwon23}. While these techniques may improve the single-device performance, in most cases, LLM inference still suffers from memory-bandwidth bottlenecks. 

\paragraph{Distributed inference}
Another major obstacle to LLM inference is the massive size of the models and KV-cache, which typically exceed the HBM capacity of a single accelerator for large models and long context sizes. To enable serving LLM inference for such large models with long context sizes, numerous prior work have focused on distributed LLM inference using a combination of parallelization techniques such as data-, tensor-~\cite{Shoeybi19}, pipeline-~\cite{Aminabadi22, fastertransformer}, expert-~\cite{Fedus22, Rajbhandari22, Singh23}, and context-parallelism~\cite{Li23}. While these techniques might reduce the memory requirements and computational workload of LLM inference per device, the overhead of communication between the devices often places an upper limit on the scalability of distributed LLM inference systems.

To mitigate the communication overhead in distributed LLM inference, prior work have proposed fused kernels that overlap computation with communication~\cite{Rashidi21, Hoefler08, Punniyamurthy24, WangWei23, Chang24}. In these methods, the compute and communication operations are partitioned into small chunks and processed in a pipeline fashion so that they overlap with each other. However, these fused kernels have several limitations. First, partitioning compute and communication operations into smaller chunks often reduces their computational efficiency due to factors such as short and irregular memory accesses. Second, this approach would require developing new kernels for each combination of compute and communication operation types, requiring significant engineering efforts. Third, the communication can be overlapped only with the immediate next compute operation, thus they do not allow overlapping multiple compute operations with the communication. For instance, these techniques can not overlap reading KV-cache with allreduce in an Attention layer because self-attention does not immediately follow allreduce in Transformer architectures. Therefore, these techniques do not provide a practical, effective, and complete solution towards hiding communication overhead in distributed LLM inference.

\paragraph{Other system-level optimizations}
Besides kernel-level optimizations and parallelization techniques, several work have proposed system-level optimizations to improve the scalability and resource utilization of LLM inference systems. Varios methods have been proposed to overlap compute-intensive prefill stage with memory-bound decode stage~\cite{Yu22, Agrawal23}. Moreover, \textit{REEF} introduces microsecond-scale kernel preemption to improve the overall throughput~\cite{Han22}. However, these methods address only memory bottlenecks and do not offer a solution to communication overheads. Alternatively, overlapping communication with compute by interleaving multiple requests has also been proposed~\cite{Du24}. Unfortunately, this approach requires splitting a batch of requests into two streams, which leads to the computational inefficiencies and resource contention in case compute and communication do not align perfectly. 

To hide the memory and I/O latencies in LLM inference systems, software prefetching techniques have been used. \textit{PrefetchML} proposes a software framework that prefetches large ML models from a database to the host memory~\cite{Daniel16}. \textit{ZeRO-Inference} prefetches a number of layers ahead of time from CPU memory or NVMe to the GPU memory to hide the latency of communication between host and devices~\cite{Aminabadi22}. Similarly, \textit{DeepUM} leverages correlation prefetching to hide the latency of page migration between GPU and CPU memories~\cite{Jung23}. \textit{RaLMSpec} introduces a speculative retrieval mechanism to predict and prefetch documents from a knowledge database to device memory in retrieval-augmented language models (RaLM) inference~\cite{Zhang24}. While these methods demonstrate the potential of software prefetching in LLM inference, none of these proposals leverage the on-chip L2 cache to maximize the data locality and do not address the challenges of limited L2 cache size for this purpose. 

Although the prior work, as mentioned above, has proposed various optimization techniques to improve the efficiency, scalability, and performance of LLM inference systems, to the best of our knowledge, none of them has proposed a mechanism that allows prefetching the model weights and KV-cache from device memory to the on-chip L2 cache while overlapping the communication with memory operations to reduce the overhead of inter-device communication. Therefore, this work proposes a novel solution to an important problem in distributed LLM inference systems.

\section{Limitations of the Proposed Method}

While the previous sections have demonstrated the potential benefits, we now discuss the known limitations of the proposed method. First, the effectiveness of the proposed method is limited by how much model weights and KV cache can fit in the on-chip memory capacity of the AI accelerators. Therefore, any factor that increases the model weight size and KV-cache per layer and per device (e.g., batch size, sequence length, precision) may affect the speedup obtained with the proposed method. However, tight latency constraints in online inference applications (e.g., chatbots, code assistants) often place an upper bound on the batch size. Moreover, for long sequence lengths, sequence parallelism is often used, which reduces KV-cache size per device. Finally, more aggressive and efficient quantization methods~\cite{Lin24, Hooper24} are being developed, which further reduces the memory requirements with minimal impact on model accuracy. Therefore, the proposed method remains effective and applicable to a wide range of application settings.

\section{Conclusions}

In this work, we introduced PRESERVE, a novel prefetching framework for model weights and KV-cache that enables to overlap HBM reads with collective communication operations, hiding the latter's latency. We shared the implementation details of PRESERVE, which uses a graph optimization algorithm to automatically insert prefetching commands in parallel to communication collectives in a computation graph while preventing cache pollution without any required modification in the user code. 

We performed a series of experiments on a commercial AI accelerator and showed that the proposed method achieves up to 1.6$\times$ end-to-end speedup on state-of-the-art LLMs. Furthermore, we conducted a design space exploration to identify the optimal hardware design for AI accelerators with PRESERVE. We showed that, the optimal L2 cache size shifts from 8 MB to 104 MB when the proposed prefetching scheme is taken into consideration. Our analysis showed a further 1.25$\times$ improvement on performance per cost with PRESERVE for AI accelerators with the optimal L2 cache size. 

We draw two important conclusions from this study. First, prefetching the model weights and KV-cache from HBM memory to L2 cache is an effective way to hide the communication overhead in distributed LLM inference systems. Second, considering prefetching during hardware design time significantly improves the performance per cost of the AI accelerators. Using the proposed method on optimally designed AI accelerators is, therefore, key to improve the performance and efficiency of the distributed LLM inference systems.

\bibliographystyle{ACM-Reference-Format}
\bibliography{biblio}

\newpage
\section*{Appendix}

Due to the page limitations, we discuss some of the experiments in this section.

\begin{figure}[t!]
\centering     
\includegraphics[width=0.48\textwidth]{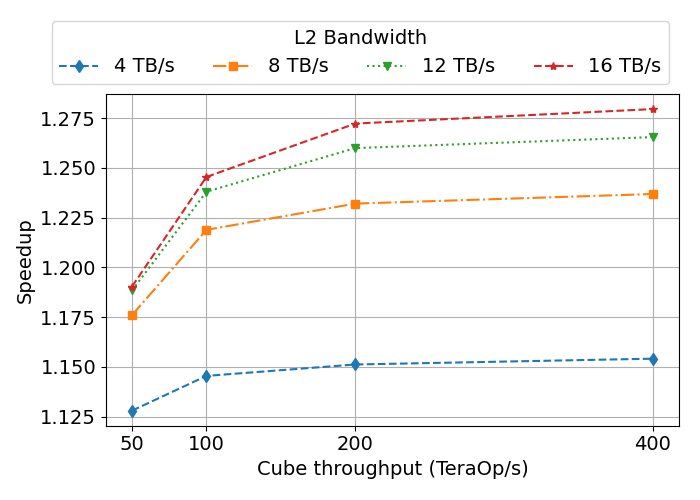}
\caption{The average speedup obtained with prefetching over the baseline with respect to the cube throughput for various L2 bandwidth values. The L2 cache for the accelerator is taken as 104 MB and the cluster size is equal to 32. The batch size and max. sequence length are equal to 16 and 8k, where prompt and decode lengths are 2/3 and 1/3 of the total number of tokens, respectively.}
\label{fig:l2-bw}
\end{figure}

\paragraph{Extending DSE to L2 bandwidth and compute units}
As the model weights and KV-cache are prefetched to the L2 cache, how quickly the prefetched data is consumed is also critical to the effectiveness of the proposed method. In most accelerators, data is fetched from off-chip memory (e.g., HBM) first to L2 cache, and then to the compute cores during the execution of an operation. As a result, the L2 latency is mostly hidden by the off-chip memory bandwidth, which is typically an order of magnitude slower than L2 bus. However, as the proposed method prefetches the data to L2 cache prior to the execution of an operation, the access latency to L2 is exposed during the computation and hinders the performance improvements obtained by the proposed method. Therefore, we now analyze the impact of the cube throughput and L2 bus bandwidth on the performance of the proposed method. 

Figure \ref{fig:l2-bw} plots the speedup obtained with prefetching over the baseline accelerator for various cube throughput and L2 bus bandwidth values. As the cube throughput increases, the speedup obtained by the proposed method also increases, as the execution time of the compute-bound operations (e.g., prefill operations) decreases. The plot also shows how the speedup versus cube throughput changes with L2 bandwidth. As expected, we also observe that increasing the L2 bandwidth improves speedup obtained by the proposed method by up to 10\%. This is because the exposed latency of the L2 access during the computation becomes shorter, improving the effectiveness of the proposed method. Therefore, we conclude that the access latency and bandwidth of the  L2 bus is an important design consideration for accelerators that adopt the proposed prefetching scheme.

\paragraph{Multi-node scale-out}
In Section \ref{sec:experiments}, we discussed the impact of the cluster size (number of accelerators) on the speedup obtained by the proposed method up to a cluster size of 8, which is the size of single Atlas 800T A2 server. In this section, we now extend this analysis beyond a single server using our accelerator performance model to evaluate the performance gains by the proposed method on a scale-out system. 

We expect that the maximum speedup occurs when the allreduce and prefetch latencies are equal to each other. If the prefetch latency is shorter than the allreduce, the latter's latency can not be fully hidden. In contrast, if the prefetch latency is longer than the allreduce, the impact of hiding the allreduce latency on the overall speedup is reduced. As a result, the maximum theoretical speedup is obtained in settings where the prefetch and allreduce latencies are equal to each other.

Increasing the number of devices in a cluster typically decreases the local memory size of model weights and KV-cache, as these tensors are partitioned into smaller chunks. Consequently, the data size that needs to be read from HBM memory decreases, reducing the prefetch time. As a result, adding or removing devices from the cluster may change the ratio between allreduce and prefetch latencies, which also affects the speedup obtained by the proposed method. Therefore, we perform an analysis on how the speedup is affected by the cluster size.

\begin{figure}[t]
\centering     
\includegraphics[width=0.48\textwidth]{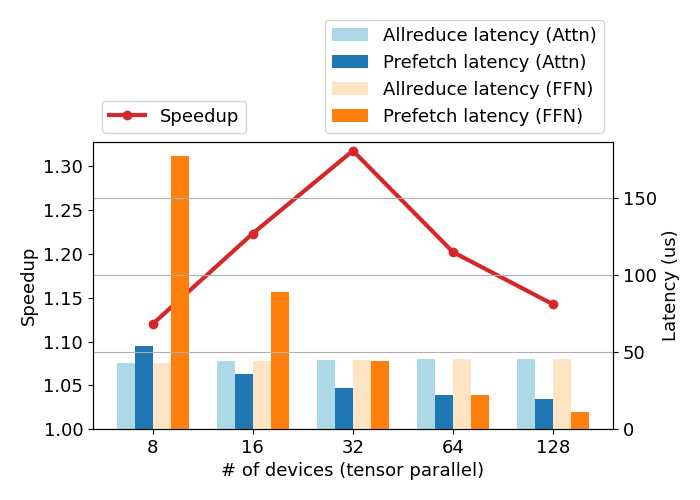}
\caption{Speedup obtained with prefetching over the baseline (left-side y-axis) as well as the allreduce and prefetch latencies for Attention and FFN layers (right-side y-axis) versus various number of (tensor-parallel) devices. The L2 cache for the accelerator is taken as 104 MB. The benchmark is a Llama3-405B model with a batch size of 16 and max. sequence length of 8k, where prompt and decode lengths are 2/3 and 1/3 of the total number of tokens, respectively.}
\label{fig:scaleout}
\end{figure}

Figure \ref{fig:scaleout} shows the speedup obtained by prefetching over the baseline accelerator for various number of devices in a tensor-parallel cluster. We observe that the speedup increases with the number of devices up to 32, beyond which it starts to decreases. Figure \ref{fig:scaleout} also shows the allreduce and prefetch latencies for Attention and MLP layers to better understand the trends in speedup with respect to the number of devices. We observe that, at the lower end of the range of number of devices (i.e., 8 and 16), the prefetch latency for FFN layers are significantly longer than the allreduce latencies, which limits the speedup. With the increasing number of devices, the weight size of FFN layers for each device decreases thanks to tensor parallelization. As a result, the prefetch latency for FFN also decreases and matches the allreduce latency at the cluster size of 32, where we observe the highest speedup. Increasing the number of devices beyond the cluster size of 32 results in prefetch latencies shorter than the allreduce latencies, in which case the latter can not be fully hidden, leading to a decrease in the speedup.

\end{document}